\documentclass[10pt,twocolumn,letterpaper]{article}

\usepackage{iccv}
\usepackage{times}
\usepackage{epsfig}
\usepackage{graphicx}
\usepackage{amsmath}
\usepackage{amssymb}
\usepackage{color,xcolor,ucs}

\usepackage{algorithm}
\usepackage{algorithmic}
\usepackage{amsfonts}
\usepackage{amsthm}  
\usepackage{wasysym}
\usepackage[ruled,vlined,algo2e]{algorithm2e}
\providecommand{\SetAlgoLined}{\SetLine}
\usepackage{color}
\usepackage{dsfont}	
\usepackage{wrapfig}
\usepackage{footnote}
\usepackage{enumitem}
\usepackage{url}

\usepackage{filecontents}

\theoremstyle{definition}

\newcommand{\reid}{Re-ID }

\iccvfinalcopy 

\def\iccvPaperID{2581} 

\ificcvfinal\fi
\begin{document}

\title{Group Membership Prediction}

\author{Ziming Zhang, Yuting Chen, and Venkatesh Saligrama\\
Department of Electrical \& Computer Engineering, Boston University\\
{\tt\small \{zzhang14, yutingch, srv\}@bu.edu}
}

\maketitle

\begin{abstract}
The group membership prediction (GMP) problem involves predicting whether or not a collection of instances share a certain semantic property. For instance, in kinship verification given a collection of images, the goal is to predict whether or not they share a {\it familial} relationship. In this context we propose a novel probability model and introduce latent {\em view-specific} and {\em view-shared} random variables to jointly account for the view-specific appearance and cross-view similarities among data instances. Our model posits that data from each view is independent conditioned on the shared variables. This postulate leads to a parametric probability model that decomposes group membership likelihood into a tensor product of data-independent parameters and data-dependent factors. We propose learning the data-independent parameters in a discriminative way with bilinear classifiers, and test our prediction algorithm on challenging visual recognition tasks such as multi-camera person re-identification and kinship verification. On most benchmark datasets, our method can significantly outperform the current state-of-the-art. 
\end{abstract}

\section{Introduction}

\begin{figure}[t]
\begin{minipage}[b]{0.49\linewidth}
 \begin{center}
 \centerline{\includegraphics[width=.9\columnwidth]{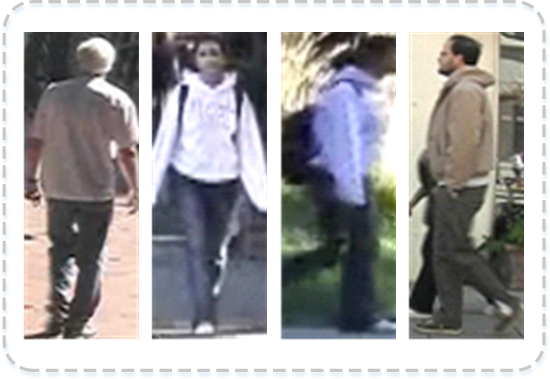}}
 \centerline{\footnotesize{(a) Person re-identification}}
 \end{center}
\end{minipage}
\begin{minipage}[b]{0.49\linewidth}
\begin{center}
\centerline{\includegraphics[width=.9\columnwidth]{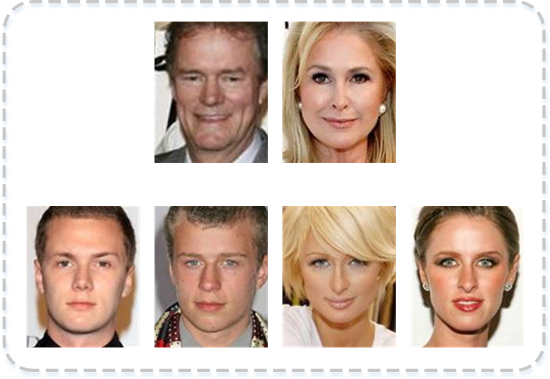}}
 \centerline{\footnotesize{(b) Kinship verification}}
\end{center} 
\end{minipage}
\vspace{-2mm}
\caption{\footnotesize{Illustration of group membership prediction (GMP) in the visual recognition tasks of (a) person re-identification and (b) kinship verification. Here we would like to predict (a) whether the four pedetrain images are taken from the same person, and (b) whether the face images are from the same family. These images are borrowed from (a) VIPeR dataset \cite{Gray_PETS07_VIPER} and (b) Family101 dataset \cite{fang2013kinship}, respectively.
}}\label{fig:model}
\end{figure}

Visual similarity plays an important role in visual recognition in object detection and scene understanding \cite{lsvm-pami,DBLP:dblp_conf/cvpr/LiS009}. A visual similarity function returns a score of how likely two instances (\eg images and videos) share similar semantic concepts (\eg persons, cars, \etc). With this perspective we propose the {\em Group Membership Prediction} (GMP) problem, where the goal is to determine how likely a collection of distinct items share the same semantic property. Fig. \ref{fig:model} depicts the idea of the GMP problem for two visual recognition tasks, \ie person re-identification and kinship verification. In person re-identification (Re-ID) we are given a collection of images of persons captured from multiple views (cameras) and the goal is to detect whether or not they belong to the {\it same} person. In applications such as kinship detection, the underlying semantic property is more general, and the goal is to predict whether or not a collection of images share a {\it familial} relationship. 
GMP poses significant challenges on account of large variations in data including lighting conditions, poses and camera views.

\begin{figure}[t]
\begin{minipage}[b]{0.49\linewidth}
 \begin{center}
 \centerline{\includegraphics[width=.9\columnwidth]{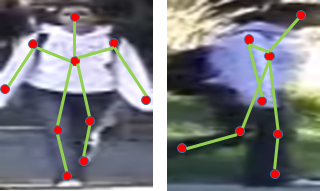}}
 \centerline{\footnotesize{(a) Body parts}}
 \end{center}
\end{minipage}
\begin{minipage}[b]{0.49\linewidth}
\begin{center}
\centerline{\includegraphics[width=.9\columnwidth]{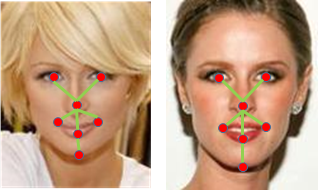}}
 \centerline{\footnotesize{(b) Facial landmarks}}
\end{center} 
\end{minipage}
\vspace{-2mm}
\caption{\footnotesize{Illustration of (a) body parts (\eg head, torso, legs) for \reid and (b) facial landmarks (\eg eyes, nose, mouth) for kinship verification. Note that in these aligned images, these body parts or facial landmarks approximately coincide in terms of spatial locations.}}\label{fig:part}
\end{figure}

We introduce a novel parametric probability model for predicting group membership. Our {\em key insight} is that although the visual appearances can significantly vary, they share a set of latent variables common to all views. As depicted in Fig. \ref{fig:part}, we can hypothesize ``body parts'' as shared latent variables for all the pedestrian images, while for kinship verification ``facial landmarks'' could be considered as the shared latent variables. Our model postulates that conditioned on the location of each shared latent variable (body part or facial landmark) the visual appearance at that location is conditionally independent for different views. This property leads to a natural way of measuring image similarities through comparison of visual similarities of the same shared latent variables across different views. 


This postulate leads us to a joint parametric probability model that consists of {\em view-specific} and {\em view-shared} random variables. View-specific variables account for visual characteristics within a view while view-shared variables account for the integration of multi-view information. The group membership likelihood factorizes into a tensor product consisting of data-independent and data-dependent factors. We learn the data-independent parameters (\ie weights) discriminatively using bilinear classifiers. Finally we marginalize these data tensors over all the dimensions with the learned weights as the group membership scores. Our experimental results on multi-camera person \reid and kinship verification demonstrate the good prediction performance and computational efficiency of our method.
 


\subsection{Related Work}
GMP problem is closely related to multi-view learning (MVL). Indeed, our perspective of shared variables has been used before in the context of MVL \cite{conf/avss/FigueiraBMCBM13,conf/cvpr/SongMD12,xu2013survey}. Nevertheless, the goal of MVL specifically in visual recognition is different from ours. Namely, the objective of MVL is to leverage multiple sources (\eg texts, images, videos, \etc) of data corresponding to the same underlying object (\eg persons, events, \etc) to improve recognition performance \cite{CaiWPQ14,Kalayeh_2014_CVPR,Liu2014,xu2013survey}. On the other hand our goal is to {\it predict} group membership among the multiple sources.


Person \reid essentially is a GMP problem, where each camera view can be taken as one of the instances. In the literature, however, most of existing works consider this problem as an independent two-view classification task, mainly focusing on cleverly designing local features \cite{Farenzena_CVPR10_SDALF,Liu_ECCV12_importance,Pedagadi_CVPR13_LFDA,yang_eccv14,zhao2014learning} or learning better metrics \cite{khamis-eccvw2014,Kostinger_CVPR12_KISSME,conf/cvpr/LiW13,conf/cvpr/LiCLHCS13,Mignon_CVPR12_PCCA,Zheng_PAMI13_RDC}. Recently, Figueira \etal \cite{conf/avss/FigueiraBMCBM13} proposed a semi-supervised learning method to fuse multi-view features for \reid so that the features agree on the classification results. Das \etal \cite{Das2014} considered the group membership prediction in \reid by maximizing the summation of pairwise similarity scores using binary integer programming during testing. Unlike \cite{Das2014}, we formulate the group membership problem as a learning problem, rather than a post-processing step to improve the matching rate.


Kinship verification is indeed another GMP problem, where each family role (\eg father, mother, son, daughter, \etc) can be considered as an instance. Similar to person Re-ID, existing works mainly focus on learning better features \cite{dehghan2014look,fang2010towards} and better distance metrics \cite{lu2014neighborhood} for pairwise classification \cite{lu2015fg}. Recently, Qin \etal \cite{qin2015tri} proposed a bilinear model to handle so-called tri-subject kinship verification problems. Fang \etal \cite{fang2013kinship} proposed a sparse group lasso based feature selection method to determine whether a query person is from a specific family. Unlike \cite{fang2013kinship,qin2015tri}, our method targets at a more general and challenging problem which can be used to predict an arbitrary number of images with a fixed structure of family roles, such as father-son, father-mother-daughter, grandfather-father-son-grandson, \etc.

%



\section{Our Method}\label{sec:method}

\subsection{Problem Setting} \label{ssec:problem_setting}

Let $\{(\mathcal{X}_m, y_m)\}_{m=1,\cdots,M}$ be a group of $M$ persons from different views, where $\forall m, \mathcal{X}_m$ denotes the $m^{th}$ person and $y_m$ denotes its label (\eg identity or family). Let $\forall n=\{1,\cdots,N_m\}, \mathbf{x}_{m,n}\in\mathcal{X}_m$ be the $n^{th}$ image for the person with $N_m$ images in total. The goal of our method is to predict the following probability as {\em group membership}:
\begin{align}
p(y_1=\cdots=y_M|\mathcal{X}_1,\cdots,\mathcal{X}_M).
\end{align}

Note that our problem setting is naturally applicable to the {\em multiple instance} cases. For example, during learning we allow multiple images to be associated with a person (\ie $\mathcal{X}_m=\{\mathbf{x}_{m,n}\}$) in person \reid and kinship verification, as in the CUHK Campus \cite{Zhao_ICCV13_salience} and Family101 \cite{fang2013kinship} datasets.

While we have motivated our approach in the context of shared latent variables (body parts or facial landmarks), this information is unavailable during the training or testing phases. Furthermore, estimating locations of body parts and facial landmarks is known to be extremely challenging \cite{bourdev2009poselets,zhu2012face}. 
Fortunately, in the context of the applications and problems that we are concerned with, the images are approximately aligned. 
In these images, foreground objects are centralized and well cropped. Currently most benchmark datasets are composed of such approximately aligned images, namely, the same body parts or facial landmarks appear roughly at similar locations. In such cases, pixel locations provide good approximation of where body parts and facial landmarks are, and we utilize this property to bypass the detection challenge, while accounting for spatial misalignments with spatial kernels. Note that the issue of {\em visual ambiguity} of the shared variables still remains in our problem. 


\subsection{Parametric GMP Model}\label{ssec:gmi}


\begin{figure}[t]
\begin{minipage}[b]{0.49\linewidth}
 \begin{center}
 \centerline{\includegraphics[width=.9\columnwidth]{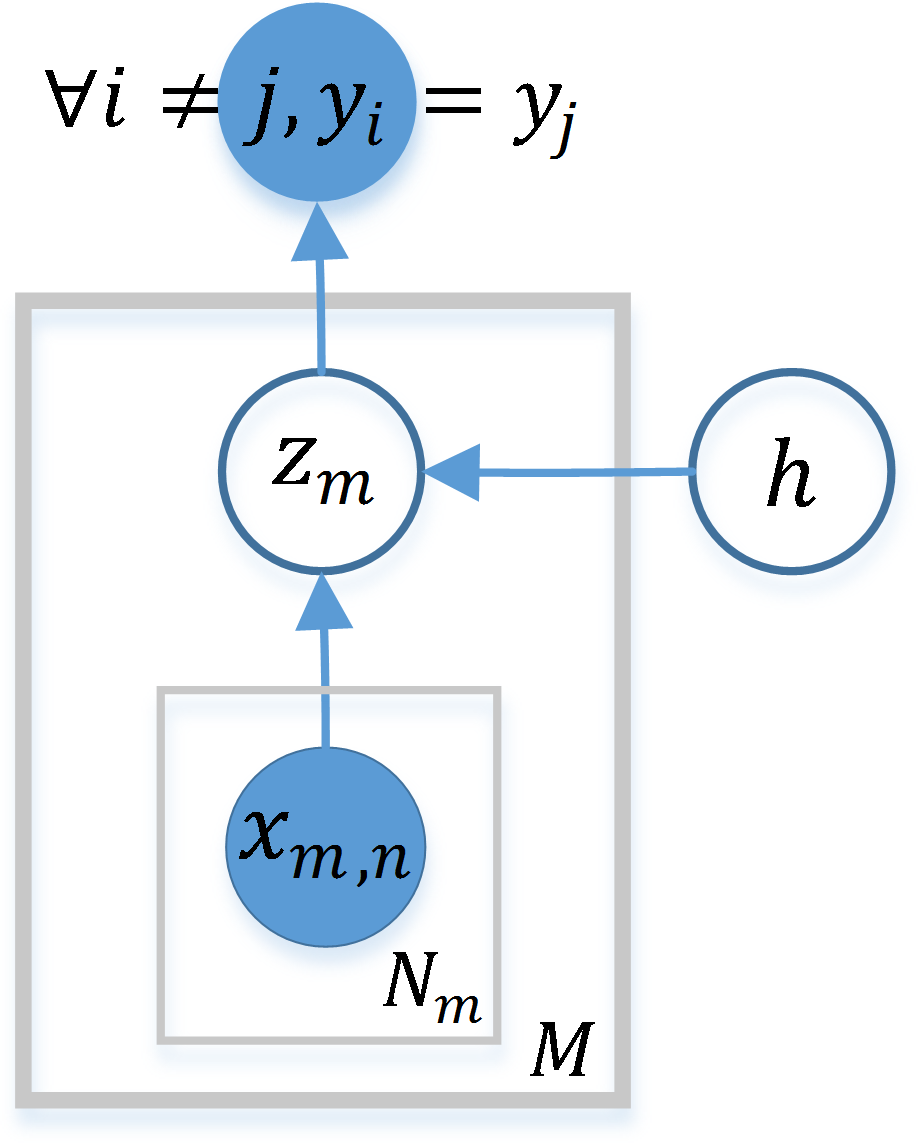}}
 \centerline{\footnotesize{(a) Parametric probability model}}
 \end{center}
\end{minipage}
\begin{minipage}[b]{0.49\linewidth}
\begin{center}
\centerline{\includegraphics[width=.7\columnwidth]{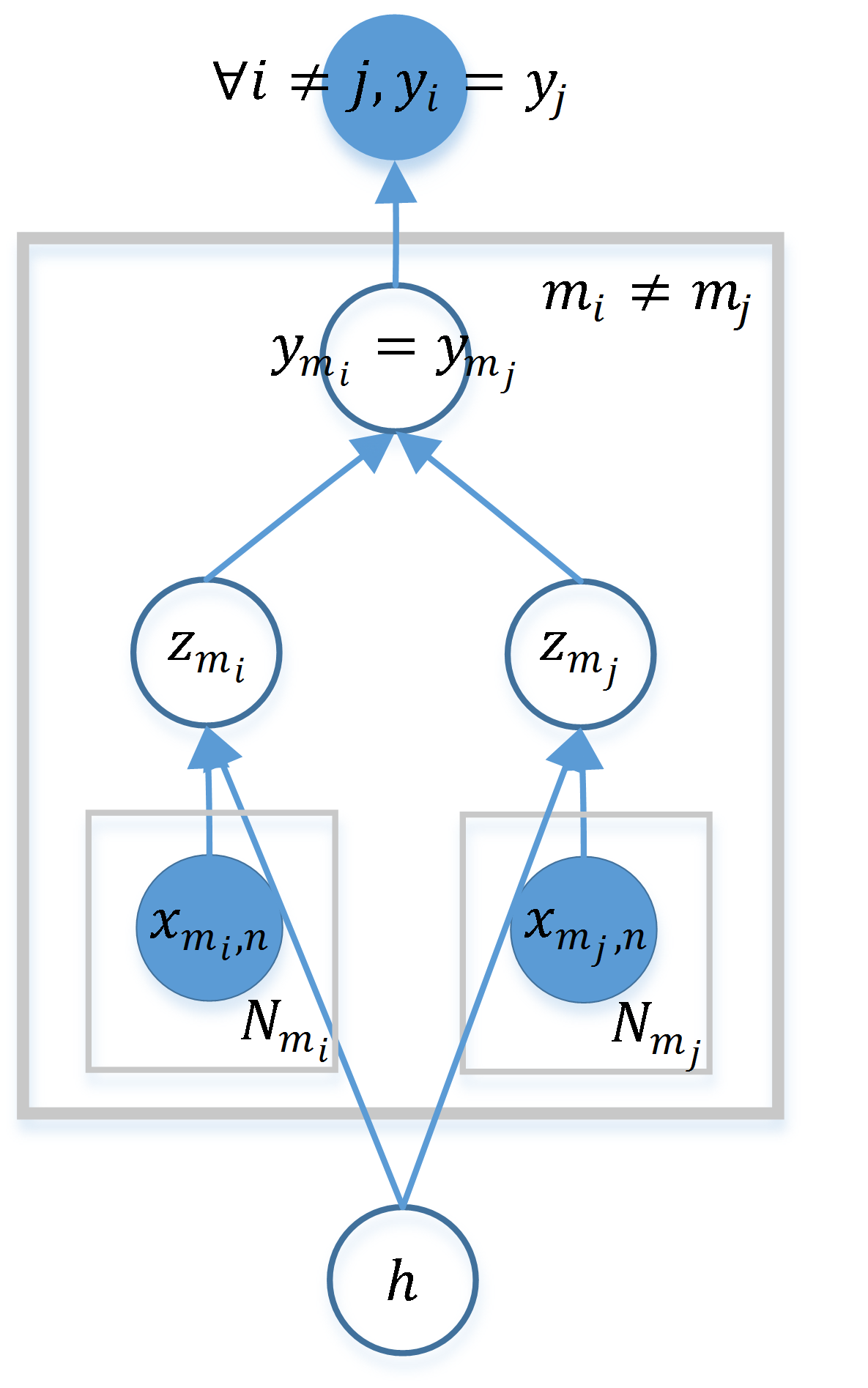}}
 \centerline{\footnotesize{(b) Pairwise decomposition}}
\end{center} 
\end{minipage}
\vspace{-2mm}
\caption{\footnotesize{(a) Graphical representation of our parametric probability model for GMP. (b) Pairwise decomposition of our model in (a).}}\label{fig:gm}
\end{figure}


We introduce two latent variables to model the relationship between the class labels $\{y_m\}$ and data samples $\{\mathcal{X}_m\}$. The graphical representation of our parametric probability model is shown in Fig. \ref{fig:gm}(a), where $\forall m, z_m$ denotes the view-specific latent variable for view $m$, $h$ denotes the view-shared latent variable, and $N_m$ denotes the number of images from view $m$. Based on this model, we can factorize our group membership score as follows:
\begin{align}\label{eqn:general}
& p(y_1=\cdots=y_M|\mathcal{X}_1,\cdots,\mathcal{X}_M)\\
& = \sum_{z_1,\cdots,z_M,h}\hspace{-2mm}p_{y|z}\prod_{m=1}^{M}p(z_m|\mathcal{X}_m,h)p(h)\nonumber\\
& = \sum_{z_1,\cdots,z_M,h}p_{y|z}\cdot p(h)\prod_{m=1}^{M}\left[\frac{1}{N_m}\sum_{n=1}^{N_m}p(z_m|\mathbf{x}_{m,n},h)\right],\nonumber
\end{align}
where $p_{y|z}=p(y_1=\cdots=y_M|z_1,\cdots,z_M)$.\\

\noindent
{\bf Model interpretation.} To show the intuition of our parametric probability model, we consider the person \reid example in Fig. \ref{fig:part}(a) in more detail. In the \reid problem the view-specific latent variables $\{z_m\}$ can be thought of as visual appearances of body parts of different persons, and the view-shared latent variable $h$ can be considered as these body parts which are shared among all the persons. 

Then using Bayes rule we can expand Eq. \ref{eqn:general}. In particular, for the two-view \reid problem we see that the group membership score of the image pair $(\mathbf{x}_1,y_1)$ and $(\mathbf{x}_2,y_2)$ as $p(y_1=y_2|\mathbf{x}_1,\mathbf{x}_2)=\sum_{z_1,z_2,h}p(y_1=y_2|z_1,z_2)p(\mathbf{x}_1|z_1,h)p(\mathbf{x}_2|z_2,h)p(h)$. Since visual appearances in $z_1$ (or $z_2$) are posited to be independent given image $\mathbf{x}_1$ (or $\mathbf{x}_2$) and the parts $h$, we can predict whether or not $y_1$ is equal to $y_2$ (\ie $p(y_1=y_2|\mathbf{x}_1,\mathbf{x}_2)$) by marginalizing the similarities of corresponding visual features of each individual part in both images (\ie $p(\mathbf{x}_1|z_1,h)$ and $p(\mathbf{x}_2|z_2,h)$) with some data-independent weights (\ie $p(y_1=y_2|z_1,z_2)$ and $p(h)$). Similarly for the kinship example in Fig. \ref{fig:part}(b) we can infer the group membership score by marginalizing the corresponding landmark similarities.
 
We take these data-independent weights as the model parameters for prediction, which are learned discriminatively.


\subsection{Discriminative Learning of Model Parameters}\label{ssec:learning}

\subsubsection{Co-occurrence Tensor Representation}

As discussed in Section \ref{ssec:problem_setting}, images are approximately aligned in the related applications. Specifically, in person Re-ID benchmarks the head is always located at the top of images, torso in the middle, and legs at the bottom. This typical structure has been exploited in designing discriminative features \cite{Farenzena_CVPR10_SDALF}. Therefore, with approximately aligned images we can bypass the problem of shared variable detection and directly utilize pixel locations as surrogates for locations of body parts or facial landmarks. Note that we can still allow small spatial misalignments by designing kernels to account for spatial distortions. 


Recently, Zhang \etal \cite{zhang2014novel} proposed an interesting feature representation to handle visual ambiguity and spatial distortion in images for person {\it re-id}. The basic idea in their method is to capture visual ambiguity using visual words, and match them at similar locations using distance transform to handle spatial distortion. This results in a visual word co-occurrence matrix for a pair of images. 

Inspired by \cite{zhang2014novel}, we propose a visual word {\em co-occurrence tensor} representation using $p(z_m|\mathbf{x}_{m,n},h)$ from multiple views to represent the group of data samples.
Their proposed Gaussian kernel \cite{zhang2014novel} is computationally cumbersome. Instead we design a truncated exponential function as the spatial kernel $\kappa$ with an arbitrary distance function inside to improve flexibility and computational efficiency.

Let $\boldsymbol{\pi}_{z_m}\in\boldsymbol{\Pi}(z_m, \mathbf{x}_{m,n})$ be a pixel location where the corresponding pixel in image $\mathbf{x}_{m,n}$ is encoded using visual word $z_m$, and $\boldsymbol{\pi}_h$ be the pixel location with index $h$. Then we define $p(z_m|\mathbf{x}_{m,n},h)$ in Eq. \ref{eqn:general} as follows:
\begin{align}\label{eqn:sp}
& p(z_m|\mathbf{x}_{m,n},h)\stackrel{\Delta}{=}\max_{\boldsymbol{\pi}_{z_m}\in\boldsymbol{\Pi}(z_m, \mathbf{x}_{m,n})}\kappa\left(\boldsymbol{\pi}_{z_m}, \boldsymbol{\pi}_h;\sigma_m\right)\\
& =\left\{
\begin{array}{ll}
\exp\left\{-\frac{\min_{\boldsymbol{\pi}_{z_m}}d(\boldsymbol{\pi}_{z_m}, \boldsymbol{\pi}_h)}{\sigma_m}\right\}, & \mbox{if} \; d(\boldsymbol{\pi}_{z_m}, \boldsymbol{\pi}_h)\leq\alpha\\
0, & \mbox{otherwise}.
\end{array}
\right.\nonumber
\end{align}
where $d(\cdot,\cdot)$ denotes a distance function, $\sigma_m\geq0$ denotes a predefined window size parameter for view $m$, and $\alpha\geq0$ is a predefined spatial scale parameter. Then if we take view-specific and view-shared latent variables as the dimensions in the tensor to represent the group of data, the entry at index $(z_1,\cdots,z_m,h)$ can be calculated as $\prod_{m=1}^{M}\left[\frac{1}{N_m}\sum_{n=1}^{N_m}p(z_m|\mathbf{x}_{m,n},h)\right]$.

\subsubsection{General Learning Formulation}
Here we introduce additional notations to simplify our exposition. Rather than directly representing a group of data samples $\mathcal{X}_{1,\cdots,M}=\{\mathcal{X}_1, \cdots, \mathcal{X}_M\}$ as a tensor, we convert it into a matrix $\phi(\mathcal{X}_{1,\cdots,M})\in\mathbb{R}^{\prod_{m=1}^{M}|z_m|\times|h|}$ with dimensions $\prod_{m=1}^{M}|z_m|$ and $|h|$, respectively, where $\forall m, |z_m|$ and $|h|$ denote the numbers of visual words for view $m$ and pixel locations in images. Further, we denote $\mathbf{w}_z\stackrel{\Delta}{=}p(y_1=\cdots=y_M|z_1,\cdots,z_M)\in\mathbb{R}^{\prod_{m=1}^M|z_m|}$ and $\mathbf{w}_h\stackrel{\Delta}{=}p(h)\in\mathbb{R}^{|h|}$ as our model parameters in the form of vectors. Then our group membership score in Eq. \ref{eqn:general} can be rewritten as a {\em decision function} $f$ as follows:
\begin{align}\label{eqn:decision}
f(\mathcal{X}_{1,\cdots,M}) = \mathbf{w}_z^T\phi(\mathcal{X}_{1,\cdots,M})\mathbf{w}_h,
\end{align}
where $(\cdot)^T$ denotes the matrix transpose operator. If $f(\mathcal{X}_{1,\cdots,M})\geq0$, we expect that all the members in the group have the same class label (and do not otherwise).

Let $\{(\mathcal{X}_{1,\cdots,M}^{(k)}, y_{1,\cdots,M}^{(k)})\}_{k=1,\cdots,N}$ be a set of $N$ training data groups from $M$ views, where $\forall k, y_{1,\cdots,M}^{(k)}=1$ if all the class labels in group $k$ are the same (and $-1$ otherwise).
Due to the specific form in Eq. \ref{eqn:decision}, we propose learning bilinear classifiers (\ie $\mathbf{w}_z$ and $\mathbf{w}_h$) for GMP inspired by \cite{conf/nips/PirsiavashRF09}, which used bilinear classifiers in a different context (binary classification):
\begin{align}\label{eqn:learning}
\min_{\mathbf{w}_z,\mathbf{w}_h}\hspace{-1mm} \frac{\lambda_1}{2}\|\mathbf{w}_z\|_2^2\hspace{-.5mm}+\hspace{-.5mm}\frac{\lambda_2}{2}\|\mathbf{w}_h\|_2^2\hspace{-.5mm}+\hspace{-1mm}\sum_{k=1}^N\ell\left(y_{1,\cdots,M}^{(k)}, f(\mathcal{X}_{1,\cdots,M}^{(k)})\right),
\end{align}
where $\ell(\cdot,\cdot)$ denotes the loss function (\eg hinge loss), $\lambda_1\geq0, \lambda_2\geq0$ are predefined regularization parameters, and $\|\cdot\|_2$ denotes the $\ell_2$-norm of a vector. 

Note that here we relax the probability constraint on $\mathbf{w}_z$ and $\mathbf{w}_h$ to real numbers so that Eq. \ref{eqn:learning} can be efficiently solved using alternating optimization. In each iteration, we fix one parameter (\ie $\mathbf{w}_z$ or $\mathbf{w}_h$) and use a standard support vector machine (SVM) solver to find the other parameter so that the objective value decreases monotonically, thus guaranteeing a local optimal solution.

\subsubsection{Pairwise Decomposition Approximation}\label{ssec:pd}

With sufficient training data, we can train a bilinear classifier directly using Eq. \ref{eqn:learning}. This training method, however, does not scale well with the number of views due to the high dimensional tensor representation, leading to serious computational and overfitting issues.

To overcome these issues, we propose an approximate pairwise decomposition method, as illustrated in Fig. \ref{fig:gm}(b), to reduce the parameter space. This is based on the conditional independence assumption in multi-view learning \cite{Blum:1998:CLU:279943.279962}. 
Accordingly, we can rewrite our group membership score in Eq. \ref{eqn:general} as follows:
\begin{align}\label{eqn:pairwise}
& p(y_1=\cdots=y_M|\mathcal{X}_1,\cdots,\mathcal{X}_M) \\
& \equiv\hspace{-2mm}\sum_{m_i\neq m_j\in\{1,\cdots,M\}}\sum_{z_{m_i},z_{m_j},h}p(y_1=\cdots=y_M|y_{m_i}=y_{m_j}) \nonumber\\
& p(y_{m_i}=y_{m_j}|z_{m_i},z_{m_j})p(z_{m_i}|\mathcal{X}_{m_i},h)p(z_{m_j}|\mathcal{X}_{m_j},h)p(h). \nonumber
\end{align}
where $p(y_1=\cdots=y_M|y_{m_i}=y_{m_j})$ indicates how importantly the pair of views $m_i$ and $m_j$ contribute to GMP. In this way, the number of parameters that need to be learned in our method is significantly reduced from $\left(\prod_{m=1}^{M}|z_m|+|h|\right)$ to $\left(\sum_{m_i\neq m_j}|z_{m_i}||z_{m_j}|+|h|\right)$. 

Let $\phi(\mathcal{X}_{m_i,m_j})\stackrel{\Delta}{=}p(z_{m_i}|\mathcal{X}_{m_i},h)p(z_{m_j}|\mathcal{X}_{m_j},h)\in\mathbb{R}^{(|z_{m_i}||z_{m_j}|)\times|h|}$ be the pairwise visual word matrix between views $m_i$ and $m_j$, where $\mathcal{X}_{m_i,m_j}=\{\mathcal{X}_{m_i},\mathcal{X}_{m_j}\}$. Also let $\mathbf{w}_{m_i,m_j}\stackrel{\Delta}{=}p(y_{m_i}=y_{m_j}|z_{m_i},z_{m_j})\in\mathbb{R}^{|z_{m_i}||z_{m_j}|}$ 
, and $\boldsymbol{\beta}\stackrel{\Delta}{=}p(y_1=\cdots=y_M|y_{m_i}=y_{m_j})\in\mathbb{R}^{|z_{m_i}||z_{m_j}|}$ 
. Then based on Eq. \ref{eqn:pairwise}, we can rewrite 
Eq. \ref{eqn:decision} as follows:
\begin{align}\label{eqn:pairwise-decision}
\tilde{f}(\mathcal{X}_{1,\cdots,M})=\sum_{m_i\neq m_j}\beta_{m_i,m_j}\mathbf{w}_{m_i,m_j}^T\phi(\mathcal{X}_{m_i,m_j})\mathbf{w}_h,
\end{align}
where $\beta_{m_i,m_j}$ denotes the entry in $\boldsymbol{\beta}$ for the view pair.
\floatname{algorithm}{Algorithm}
\begin{algorithm}[t]\small
\SetAlgoLined
\SetKwInOut{Input}{Input}\SetKwInOut{Output}{Output}
\Input{$\{ \phi(\mathcal{X}_{m_i,m_j})  \}_{\forall m_i\neq m_j \in \{1,...,M \} }$, $\{y_{m_i}\}_{\forall m_i \in {1,...,M}}$, $\lambda_1, \lambda_2, \lambda_3 \geq0$}
\Output{$ \{ \mathbf{w}_{m_i, m_j} \}_{m_i \neq m_j \in \{1,  ..., M\} },\mathbf{w}_h,\boldsymbol{\beta}$}
\BlankLine
Initialize $\boldsymbol{\beta}\leftarrow\mathbf{1}$, $\mathbf{w}_h \leftarrow \mathbf{1}$, $ \mathbf{w}_{m_i, m_j} \leftarrow \mathbf{1} $;\\
\Repeat{Converge}
{ 
Solve $\{ \mathbf{w}_{m_i, m_j} \} $ in Eq. \ref{eqn:multi-view} (multi-view training) or Eq. \ref{eqn:double-view} (double-view training) by fixing $\boldsymbol{\beta}$ and $\mathbf{w}_h$;\\
Solve $\mathbf{w}_h$ in Eq. \ref{eqn:multi-view} or Eq. \ref{eqn:double-view} by fixing $\boldsymbol{\beta}$ and $\{ \mathbf{w}_{m_i, m_j} \} $ ;\\
Solve $\boldsymbol{\beta}$ in Eq. \ref{eqn:multi-view} or Eq. \ref{eqn:double-view} by fixing $\{ \mathbf{w}_{m_i, m_j} \} $ and $\mathbf{w}_h$.
}
\Return $ \{ \mathbf{w}_{m_i, m_j} \}_{m_i \neq m_j \in \{1,  ..., M\} },\mathbf{w}_h,\boldsymbol{\beta}$\;
\caption{Pairwise decomposition based learning}\label{alg:decomposition}
\end{algorithm}

To learn our model parameters in Eq. \ref{eqn:pairwise-decision}, we propose two learning methods as follows, namely, {\em multi-view training} and {\em double-view training}:
\begin{align} 
& \mbox{\bf Multi-view training:} \nonumber\\
& \min_{\substack{\{\mathbf{w}_{m_i,m_j}\},\\ \mathbf{w}_h,\boldsymbol{\beta}}} \frac{\lambda_1}{2}\sum_{m_i\neq m_j}\|\mathbf{w}_{m_i,m_j}\|_2^2+\frac{\lambda_2}{2}\|\mathbf{w}_h\|_2^2+\frac{\lambda_3}{2}\|\boldsymbol{\beta}\|_2^2 \nonumber \\
& + \sum_{k=1}^N\ell\left(y_{1,\cdots,M}^{(k)}, \tilde{f}(\mathcal{X}_{1,\cdots,M}^{(k)})\right), \;\; \mbox{s.t.} \; \boldsymbol{\beta}\geq\mathbf{0}, \label{eqn:multi-view}\\
& \mbox{\bf Double-view training:} \nonumber\\
& \min_{\substack{\{\mathbf{w}_{m_i,m_j}\},\\ \mathbf{w}_h,\boldsymbol{\beta}}} \frac{\lambda_1}{2}\sum_{m_i\neq m_j}\|\mathbf{w}_{m_i,m_j}\|_2^2+\frac{\lambda_2}{2}\|\mathbf{w}_h\|_2^2+\frac{\lambda_3}{2}\|\boldsymbol{\beta}\|_2^2 \nonumber \\
& + \sum_{k=1}^N\sum_{m_i\neq m_j}\ell\left(y_{m_i,m_j}^{(k)}, \tilde{f}(\mathcal{X}_{m_i,m_j}^{(k)})\right), \;\; \mbox{s.t.} \; \boldsymbol{\beta}\geq\mathbf{0}, \label{eqn:double-view}
\end{align}
where $\forall k, y_{m_i,m_j}^{(k)}=1$ if in group $k$ the labels of the two persons $y_{m_i}=y_{m_j}$ holds; otherwise, 0. Here, $\geq$ denotes an element-wise $\geq$ operator. Both training can be done using alternating optimization with a standard SVM solver. Still local optima are guaranteed. For two-view scenarios, both training methods are essentially identical, and scale quadratically with the number of views, in general. Linear scalability is also possible if we organize all the views as cycle graphs. Difference in these two training methods comes from the loss functions, where in multi-view training $\ell$ measures the group (\ie multi-view) loss, while in double-view training $\ell$ measures the pair-view loss. Our algorithm is summarized in Alg. \ref{alg:decomposition}.
%
%
%


\begin{figure*}[t]
\begin{minipage}[b]{0.49\linewidth}
 \begin{center}
 \centerline{\includegraphics[width=0.8\columnwidth]{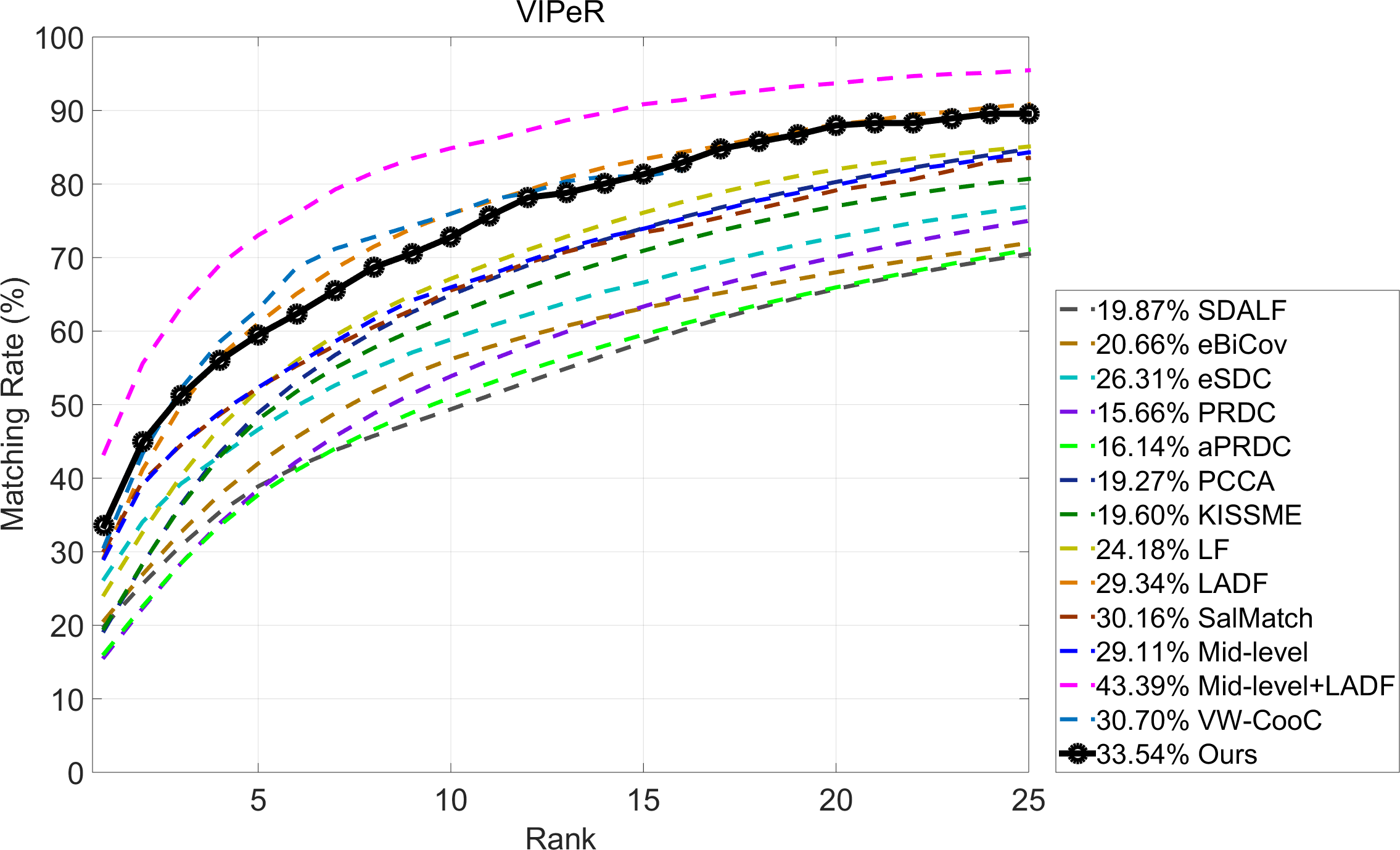}}
 \centerline{\footnotesize{(a)}}
 \end{center}
\end{minipage}
\begin{minipage}[b]{0.49\linewidth}
\begin{center}
\centerline{\includegraphics[width=0.8\columnwidth]{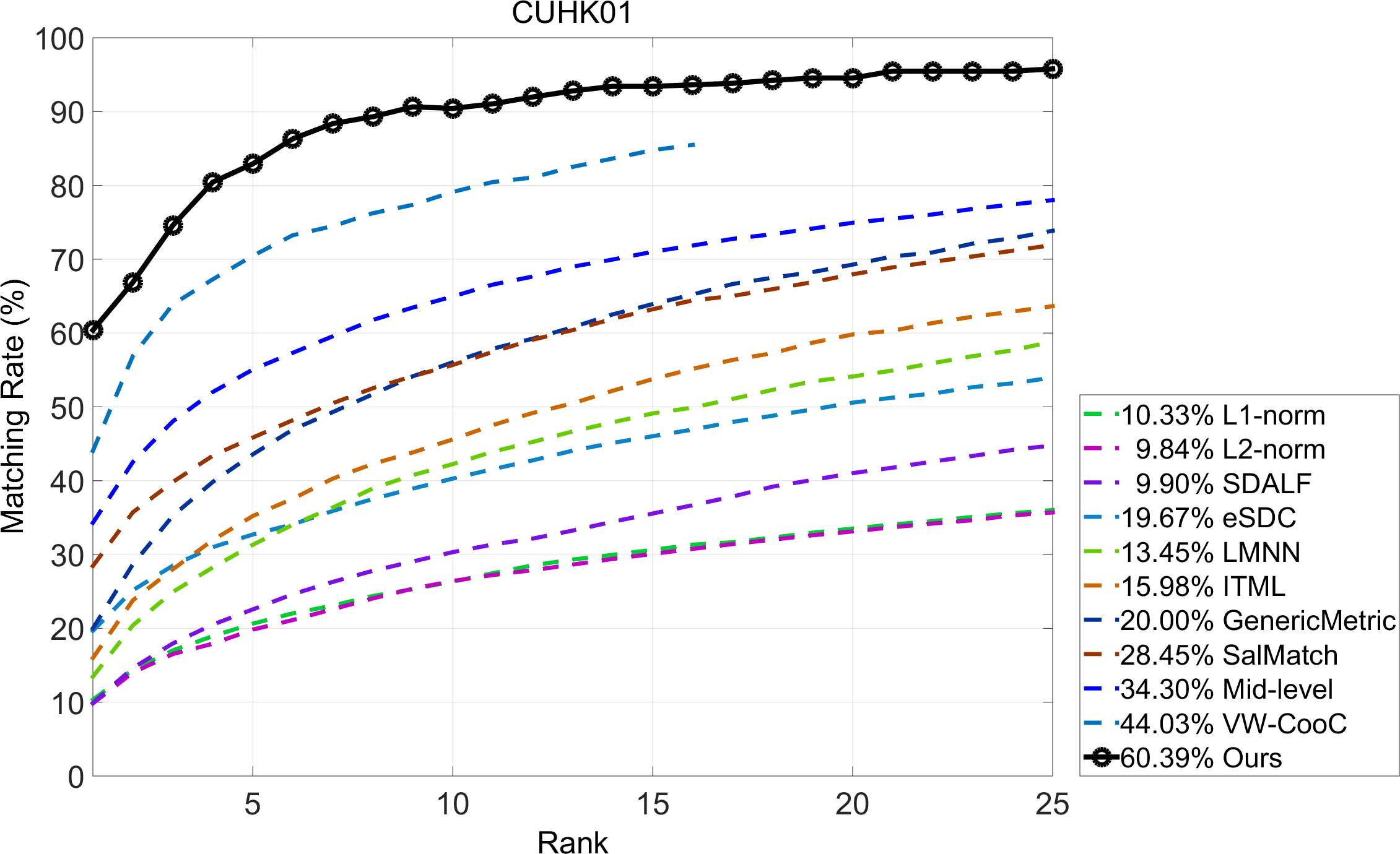}}
 \centerline{\footnotesize{(b)}}
\end{center} 
\end{minipage}
\vspace{-5mm}
\caption{\footnotesize{CMC curve comparison on \textbf{(a)} VIPeR and \textbf{(b)} CUHK01, respectively. Notice that except our results, the rest are copied from \cite{zhao2014learning}.} }\label{fig:single-shot}
\vspace{-3mm}
\end{figure*}

\section{Experiments}\label{sec:exp}

We evaluate our method on person \reid and kinship verification tasks along with state-of-the-art methods on benchmark datasets. Standard training/testing protocols are used in all experiments. For each comparing method, we either cite the original results from the papers (denoted by $(\cdot)^*$ in the tables) or calculate from released codes. Our results are reported as the average over 3 trials.






For each experiment, we choose the same or similar low-level feature as the other methods (see the details in subsection) for fair comparison. We densely sample the images to generate a low-level local feature per pixel. Then we use K-Means to build the visual vocabularies with about $2\times10^4$ randomly selected features per view. Further, every local feature is quantized into one of these visual words based on Euclidean distance. Note that more complicated feature selection methods may be employed to yield better performance, but we do not fine-tune this component for the sake of computational efficiency and generalization ability.

We employ the chessboard distance for Eq. \ref{eqn:sp} and LIBLINEAR \cite{REF08a} as our SVM solver with hinge loss. We randomly generate about $3\times10^4$ training samples to learn model parameters $\mathbf{w}$'s. The regularization parameters are determined by cross-validation.

\subsection{Person Re-identification}\label{ssec:reid}
For performance measure we adopt the standard Cumulative Match Characteristic (CMC) curve, which displays the recognition rate as a function of rank. The recognition rate at rank-$r$ is the proportion of queries correctly matched to a corresponding gallery entity at rank-$r$ or better.


For tasks with multiple camera views, we follow \cite{Das2014} to compare results under two camera views. Consider the results from multiple views as a high dimensional tensor, one dimension per view. To predict pairwise matches from multi-view results (\eg identifying matches between camera view 1 and view 2 from the predicted results for the joint of view 1, 2, and 3), we can either sum over or find the maximum over the extra dimensions. Cross-validation is used to choose the better way for each dataset.


\subsubsection{Two Camera Views}\label{ssec:two}
\begin{savenotes}
\begin{table}[t]\scriptsize\centering
\caption{\footnotesize{Matching rate comparison (\%) on VIPeR and CUHK01.}}\label{tab:ss}
\vspace{1mm}
\setlength\tabcolsep{5pt}
\begin{tabular}{|l|llllll|}
\hline
Rank $r=$ & 1 & 5 & 10 & 15 & 20 & 25 \\
\hline\hline

& \multicolumn{6}{c|}{VIPeR} \\
\hline
SCNCD \cite{yang_eccv14}    & 20.7 & 47.2  &  60.6  &  68.8  &  75.1  &  79.1  \\
SCNCD$_{final}$ \cite{yang_eccv14}  &  37.8 &  68.5 &  81.2  &  87.0  &  90.4  &  92.7  \\
LADF \cite{conf/cvpr/LiCLHCS13}    & 29.3  & 61.0  &  76.0  &  83.4  &  88.1  &  90.9 \\
Mid-level filters \cite{zhao2014learning}    & 29.1  & 52.3  &  65.9  &  73.9  &  79.9  &  84.3 \\
Mid-level+LADF \cite{zhao2014learning}    & {\bf 43.4}  & {\bf 73.0}  &  {\bf 84.9}  &  {\bf 90.9}  &  {\bf 93.7}  &  {\bf 95.5} \\
VW-CooC \cite{zhang2014novel} & 30.70 & 62.98 & 75.95 & 81.01 & - & - \\
\hline
\textbf{Ours} & 33.5 & 59.5 & 72.8 & 81.3 & 88.0 & 89.6 \\
\hline\hline
& \multicolumn{6}{c|}{CUHK01} \\
\hline
Single-shot LAFT$^*$ \cite{conf/cvpr/LiW13} & 25.8 & 55.0 & 66.7 & 73.8 & 79.0 & 83.0 \\
Multi-shot LAFT$^*$ \cite{conf/cvpr/LiW13} & 31.4 & 58.0 & 68.3 & 74.0 & 79.0 & 83.0 \\
Mid-level filters \cite{zhao2014learning}    & 34.3  & 55.1  &  65.0  &  71.0  &  74.9  &  78.0 \\
VW-CooC \cite{zhang2014novel} & 44.03 & 70.47 & 79.12 & 84.77 & - & - \\
\hline
\textbf{Ours} & {\bf 60.39} & {\bf 82.92} & {\bf 90.43} & {\bf 93.42} & {\bf 94.55} & {\bf 95.78} \\
\hline
\end{tabular}
\vspace{-3mm}
\end{table}
\end{savenotes}

Person \reid between two views is the simplest scenario. We test our method on the VIPeR \cite{Gray_PETS07_VIPER} and CUHK Campus \cite{Zhao_ICCV13_salience} dataset.  We extract a 672-dim Color+SIFT\footnote{We downloaded the code from \url{https://github.com/Robert0812/salience_match}.} vector from each 5$\times$5 pixel patch in images as low-level features.
%
We follow the experimental setting in \cite{Zhao_ICCV13_salience} for both datasets.

Our comparison results are listed in Table \ref{tab:ss}. As we see, on VIPeR ``Mid-level+LADF'' from \cite{zhao2014learning} is the current best method, which utilized more discriminative mid-level filters as features and a powerful classifier, and ``SCNCD$_{final}$'' from \cite{yang_eccv14} is the second, which utilized only foreground features. Our results are comparable to both of them. However, our method always outperforms their original methods significantly when either the powerful classifier or the foreground information is not involved. On CUHK01, our method performs the best. At rank-1, it outperforms \cite{zhang2014novel, zhao2014learning} by 16.36\% and 26.09\%, respectively. Compared with \cite{zhang2014novel}, the improvement mainly comes from the multiple instance setting of our method.

The CMC curve comparison on VIPeR and CUHK01 is shown in Fig. \ref{fig:single-shot}. As we see, our curve is very similar to that of LADF. This is mainly because LADF is a second-order (\ie quadratic) decision function based on metric learning, which shares some commonality with our classifiers.

\begin{figure}[t]\vspace{0mm}
\centerline{\includegraphics[width=0.8\columnwidth]{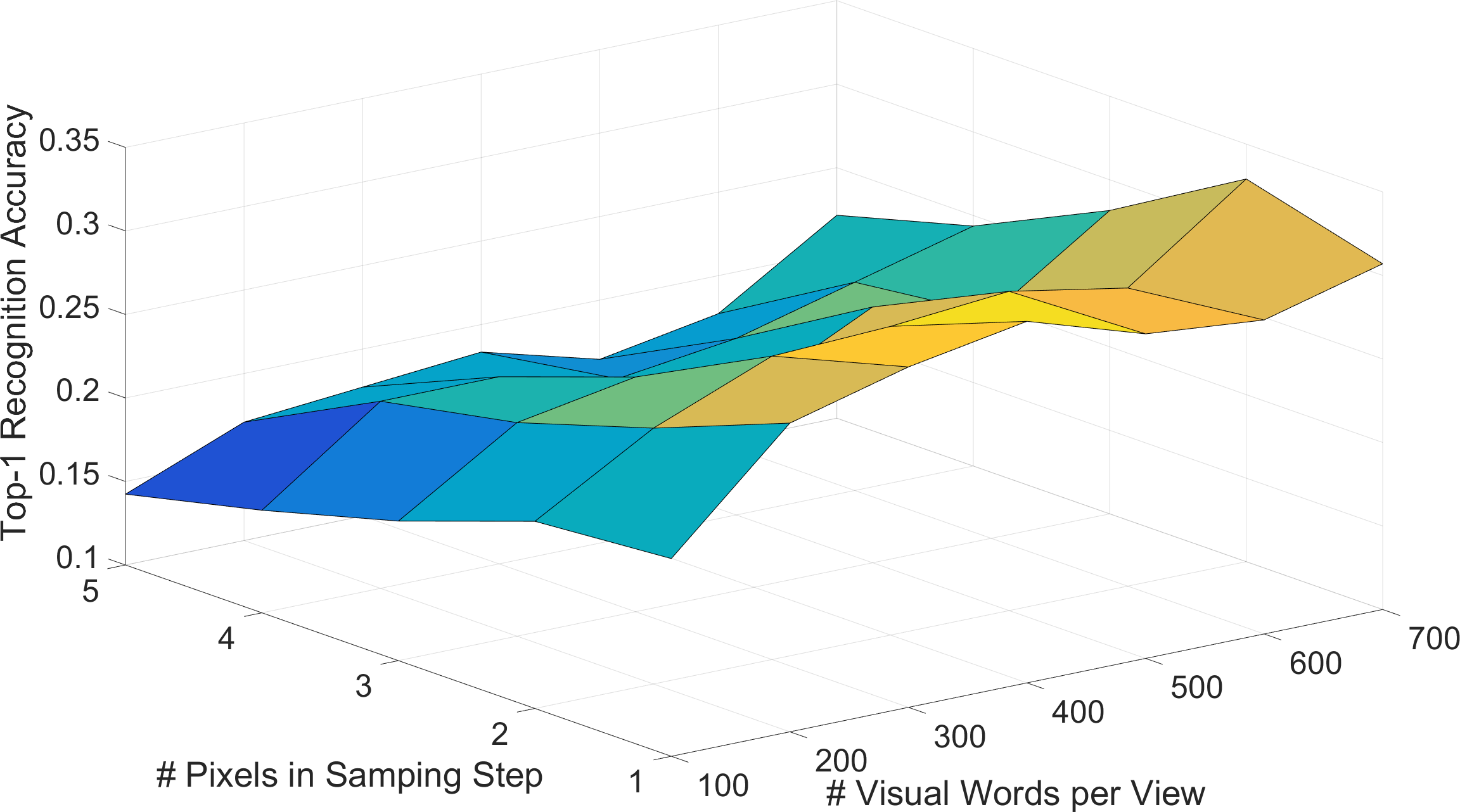}}
\caption{\footnotesize{Demonstration of the impacts of different numbers of pixel locations and visual words on the performance using VIPeR. Warmer color demotes higher accuracy. This figure is best viewed in color.}}\label{fig:loc-vw}
\vspace{-4mm}
\end{figure}

We also demonstrate the impacts of different numbers of pixel locations (\ie view-shared space) and visual words (\ie view-specific space) on the performance using VIPeR in Fig. \ref{fig:loc-vw}. We sample the pixel locations, step by from 1 to 5 pixels along x and y-axis in images (larger number leading to fewer samples), while using different numbers of visual words. Visual words capture the variations in appearance, and with more visual words more similar patterns can be differentiated (\eg pink and red). Matching between pixel locations gives us the statistic information of visual words, and more samples make the statistics more robust. Together they work for good performance.

\begin{figure*}[t]
\begin{minipage}[b]{0.33\linewidth}
 \begin{center}
 \centerline{\includegraphics[width=.72\columnwidth]{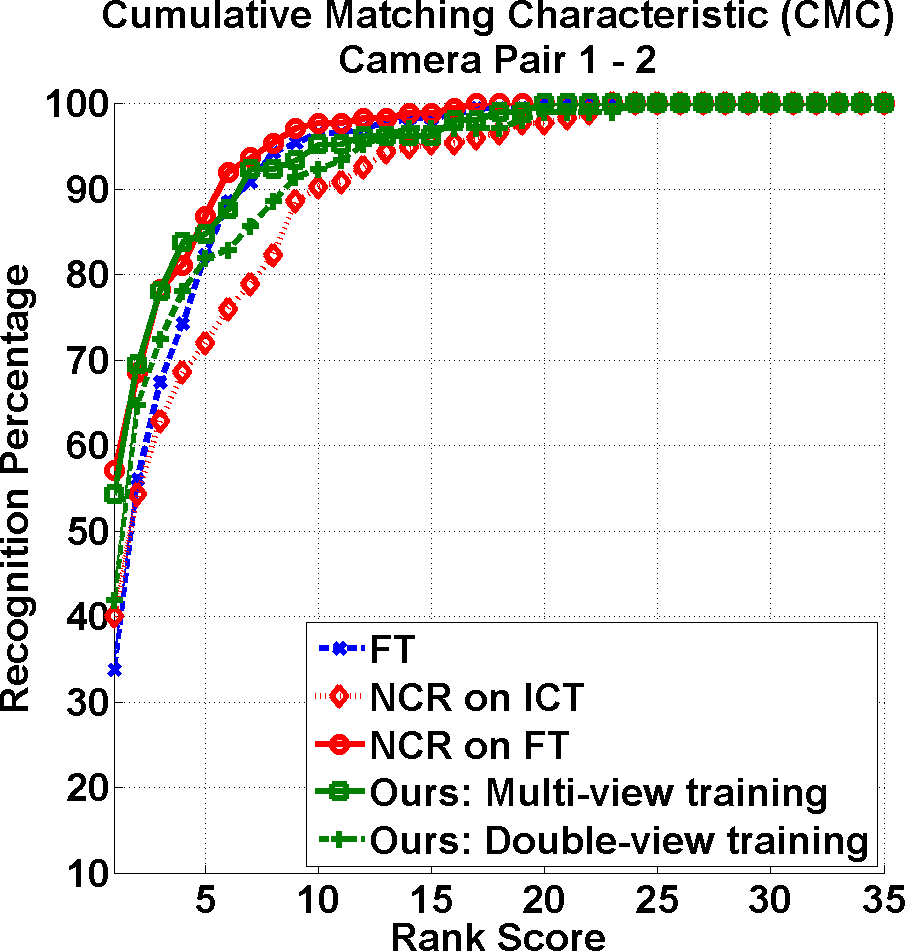}}
 \end{center}
\end{minipage}
\begin{minipage}[b]{0.33\linewidth}
\begin{center}
\centerline{\includegraphics[width=.72\columnwidth]{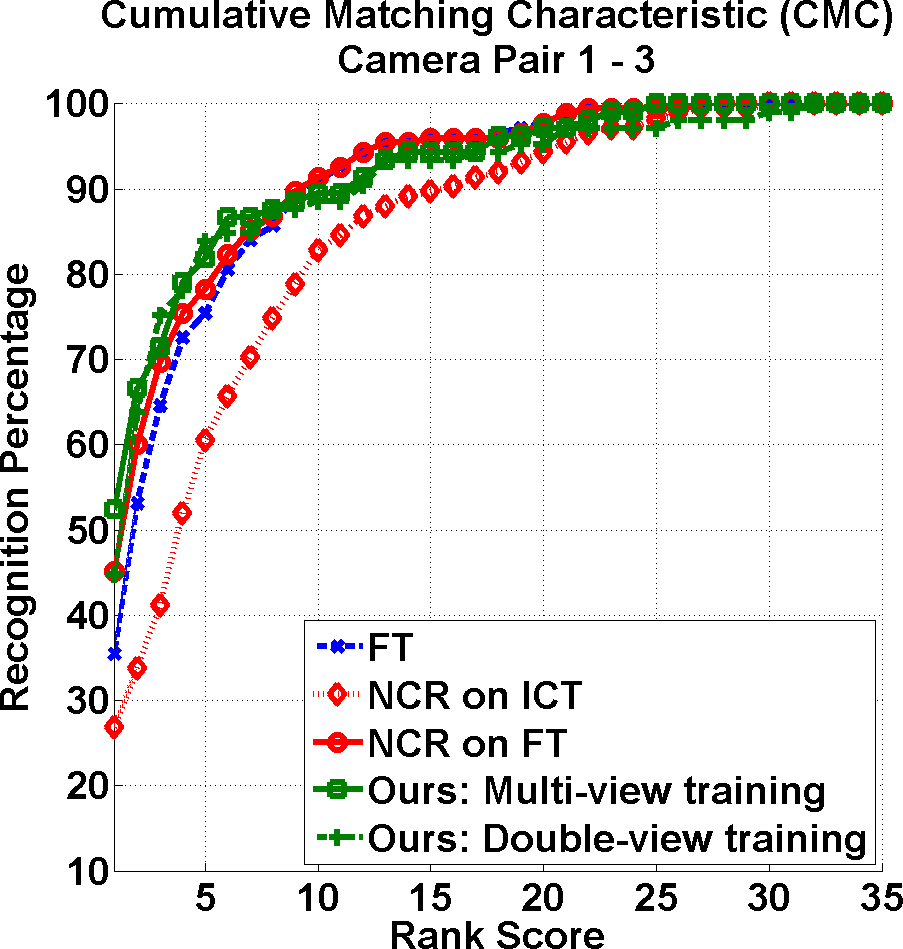}}
\end{center} 
\end{minipage}
\begin{minipage}[b]{0.33\linewidth}
\begin{center}
\centerline{\includegraphics[width=.72\columnwidth]{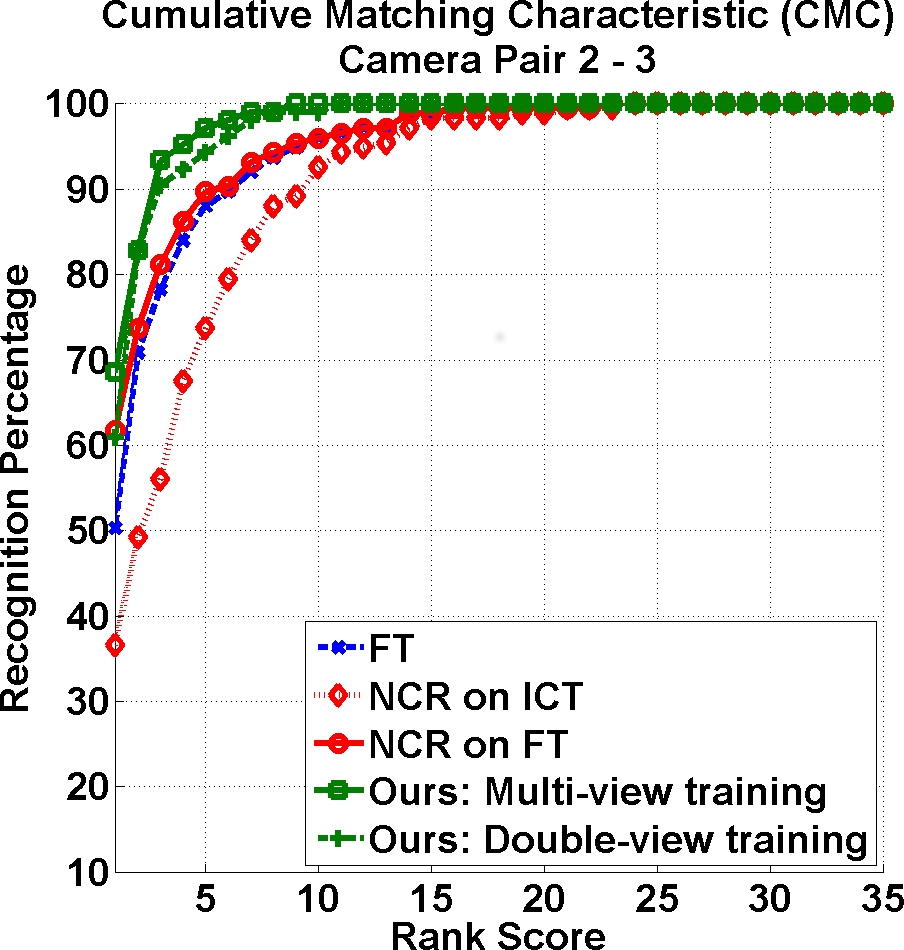}}
\end{center} 
\end{minipage}
\vspace{-7mm}
\caption{\footnotesize{CMC curve comparison on WARD. Note except for our results, the other results are cited from \cite{Das2014}.} }\label{fig:3v}
\end{figure*}

\subsubsection{Three Camera Views}\label{ssec:three}
Now we consider three camera views, and test our method on the WARD dataset \cite{conf/cvpr/MartinelM12}.
Following \cite{Das2014}, we denote the camera views as view 1, 2 and 3. However, for pairwise view matching, \cite{Das2014} did not mention which view as probe or gallery. Here, we define the view with a smaller/larger number of data to be the gallery/probe set. We randomly select 35 people for training, and the rest for testing.

We first resize each image to the same $128\times 64$ pixels, and take every $2\times2$ pixel patch in the HSV color space to generate our low-level features by concatenating $3\times2\times2=12$ entries into a vector. The reason for choosing this feature is because in \cite{Das2014} the features were built in the HSV color space as well. Different from \cite{Das2014}, we take the whole image to generate features without foreground segmentation.

The results are shown in Fig. \ref{fig:3v}. As we see, our method performs similar or better than NCR \cite{Das2014}, and the curves of both the multi-view training and double-view training for our method behave very similarly. 
We list the area under curve (AUC) scores in Table \ref{tab:3v}. Our method is better than NCR on FT by 0.6\%, on average, from 94.3\% to 94.9\%.

\begin{savenotes}
\begin{table}[t]\centering\footnotesize
\caption{\footnotesize{AUC comparison (\%) on WARD based on Fig. \ref{fig:3v}.}}\label{tab:3v}
\vspace{1mm}
\begin{tabular}{|l|lll|l|}
\hline
View pair & 1-2 & 1-3 & 2-3 & Ave.\\
\hline
FT & 93.3 & 91.0 & 94.9 & 93.1 \\
NCR on ICT & 90.4 & 84.8 & 91.1 & 88.7 \\
NCR on FT & {\bf 95.4} & {\bf 91.9} & {\bf 95.6} & {\bf 94.3} \\
\hline
{\bf Ours: Multi-view} & {\bf 94.4} & {\bf 92.1} & {\bf 98.1} & {\bf 94.9} \\
{\bf Ours: Double-view} & 92.7 & 91.0 & 97.5 & 93.8 \\
\hline
\end{tabular}
\vspace{-2mm}
\end{table}
\end{savenotes}

\begin{figure*}[t]
\begin{minipage}[b]{0.33\linewidth}
 \begin{center}
 \centerline{\includegraphics[width=.72\columnwidth]{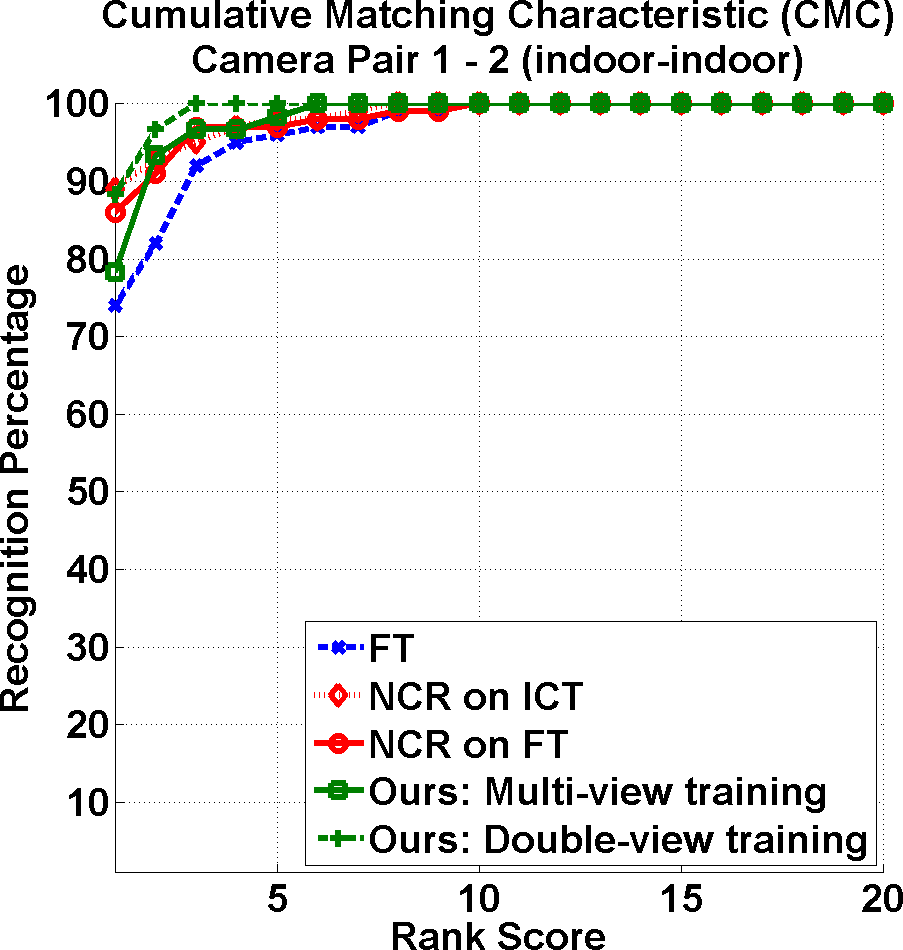}}
 \end{center}
\end{minipage}
\begin{minipage}[b]{0.33\linewidth}
\begin{center}
\centerline{\includegraphics[width=.72\columnwidth]{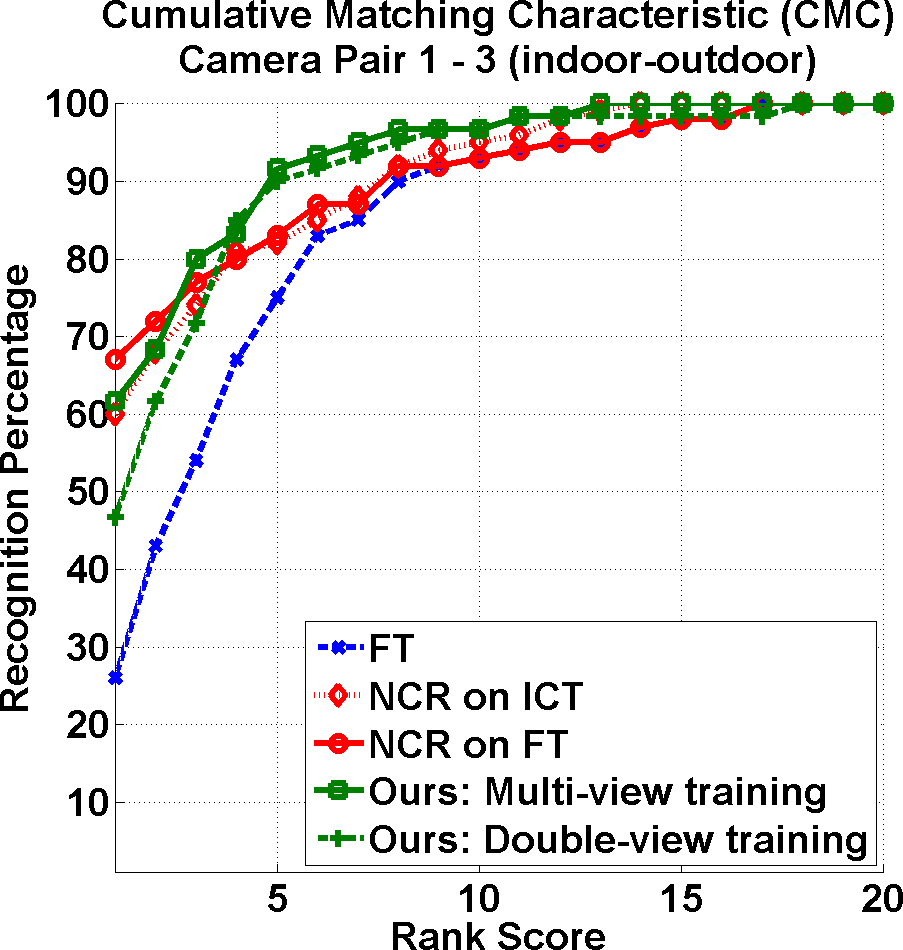}}
\end{center} 
\end{minipage}
\begin{minipage}[b]{0.33\linewidth}
\begin{center}
\centerline{\includegraphics[width=.72\columnwidth]{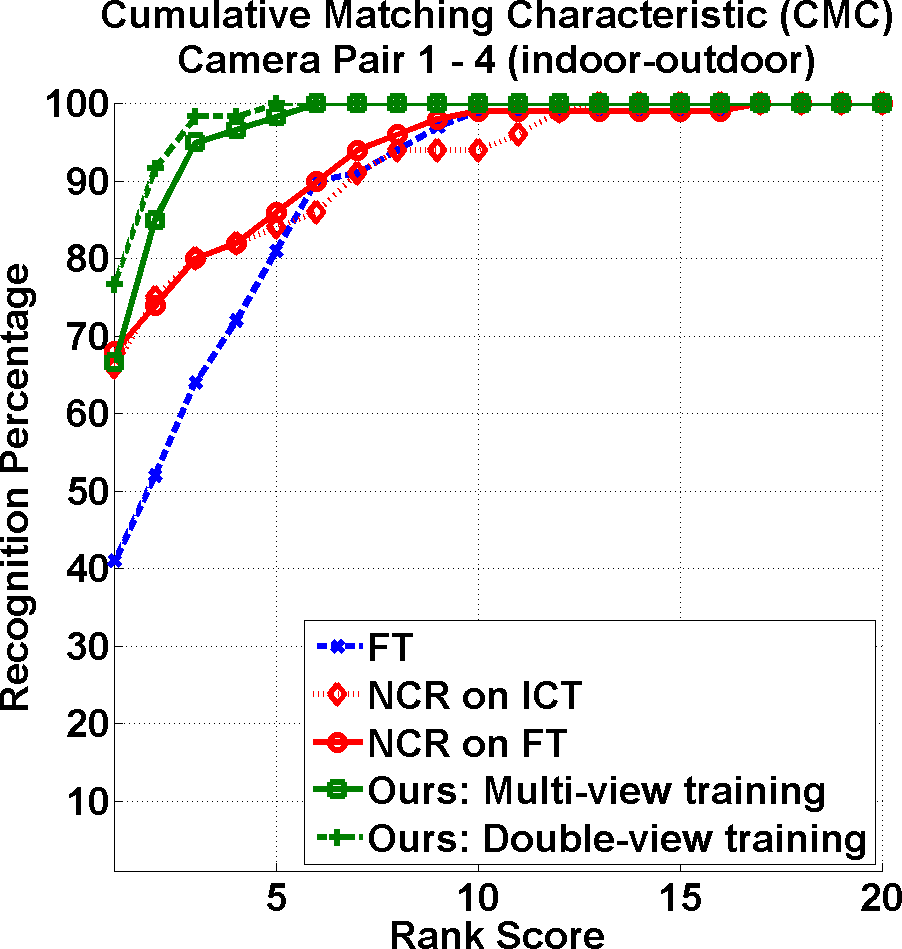}}
\end{center} 
\end{minipage}
\begin{minipage}[b]{0.33\linewidth}
 \begin{center}
 \centerline{\includegraphics[width=.72\columnwidth]{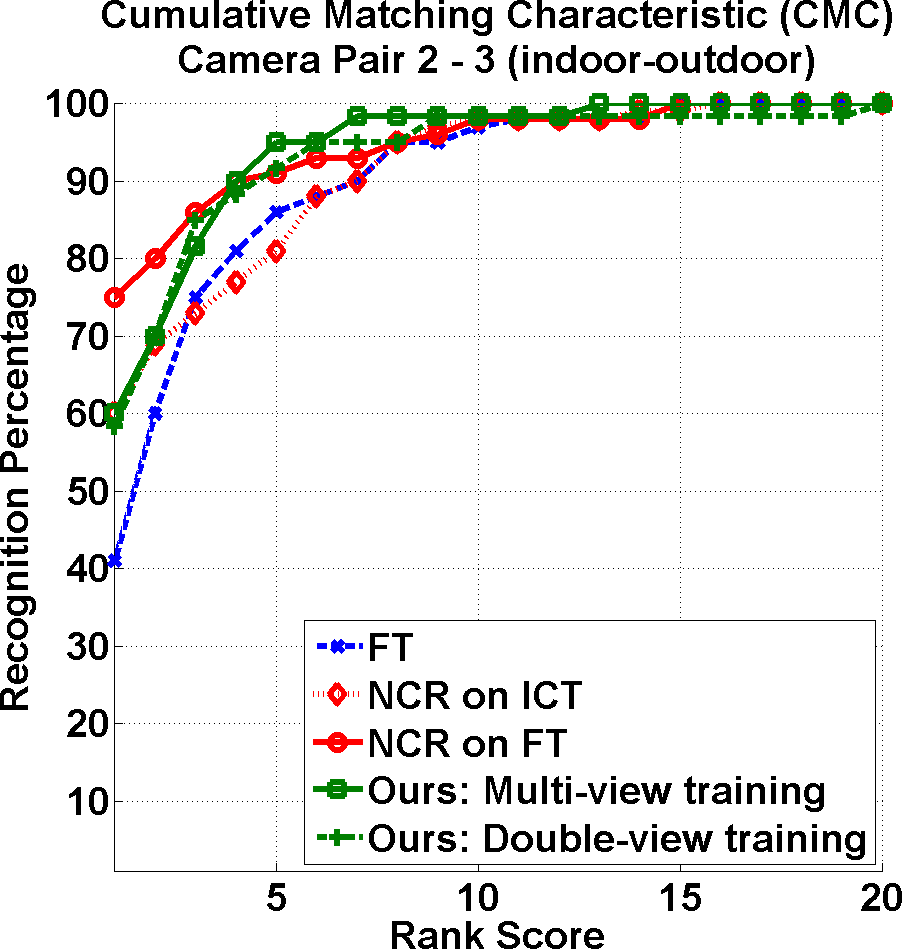}}
 \end{center}
\end{minipage}
\begin{minipage}[b]{0.33\linewidth}
\begin{center}
\centerline{\includegraphics[width=.72\columnwidth]{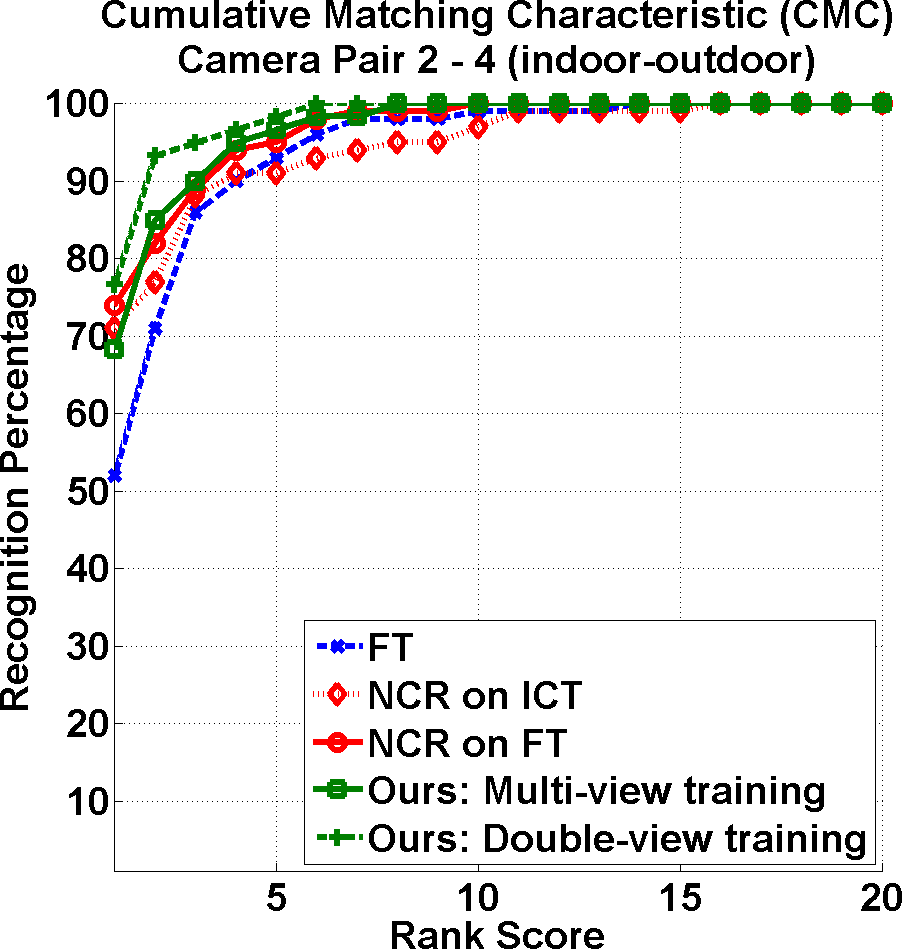}}
\end{center} 
\end{minipage}
\begin{minipage}[b]{0.33\linewidth}
\begin{center}
\centerline{\includegraphics[width=.72\columnwidth]{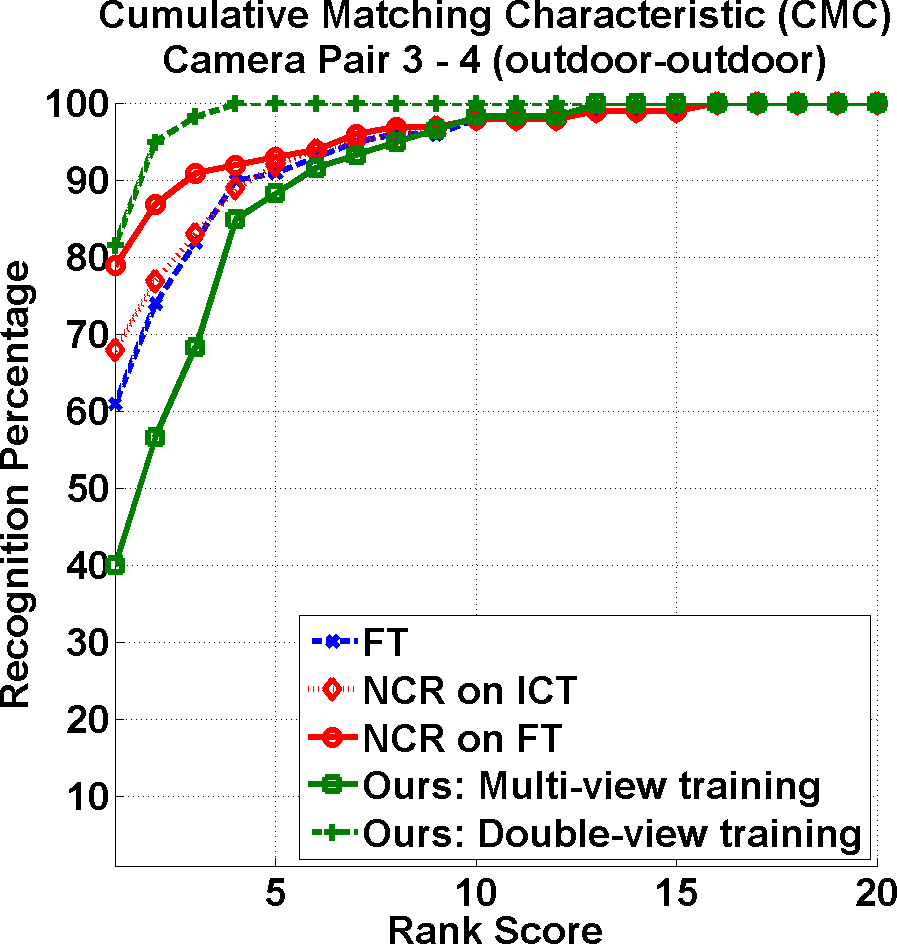}}
\end{center} 
\end{minipage}
\vspace{-8mm}
\caption{\footnotesize{CMC curve comparison on RAiD. Notice that expect our results, the rest results are cited from \cite{Das2014}.} }\label{fig:4v}
\vspace{-3mm}
\end{figure*}

\subsubsection{Four Camera Views}\label{ssec:four}
Next we consider four camera views, and test our method on the Re-identification Across indoor-outdoor Dataset (RAiD) \cite{Das2014} with two indoor views camera 1 and 2, and two outdoor views camera 3 and 4.
Still we take the views with smaller/larger numbers as galleries/probes. We follow \cite{Das2014}, 
and utilize the same HSV low-level feature as we did in Section \ref{ssec:three}.

Our comparison results are shown in Fig. \ref{fig:4v}. As we see, our method again performs equally well or better than NCR. We list the AUC score comparison results in Table \ref{tab:4v}. Still our method is better than NCR on FT by 1.6\%, on average, from 94.7\% to 96.3\%. 

For both indoor-indoor and outdoor-outdoor cases, our method consistently works best, which may indicate that the visual word co-occurrence patterns are more discriminative if the lighting condition is similar.

\begin{savenotes}
\begin{table}[t]\scriptsize\centering
\caption{\footnotesize{AUC comparison (\%) on RAiD based on Fig. \ref{fig:4v}.}}\label{tab:4v}
\vspace{1mm}
\begin{tabular}{|l|llllll|l|}
\hline
View pair & 1-2 & 1-3 & 1-4 & 2-3 & 2-4 & 3-4 & Ave.\\
\hline
FT & 96.6 & 84.3 & 88.8 & 90.0 & 93.9 & 93.5 & 91.2 \\
NCR on ICT & {\bf 98.5} & {\bf 90.6} & 92.1 & 91.0 & 94.4 & 94.1 & 93.4 \\
NCR on FT & 98.1 & 90.4 & {\bf 93.1} & {\bf 94.5} & {\bf 96.5} & {\bf 95.9} & {\bf 94.7} \\
\hline
{\bf Ours: Multi-view} & 98.2 & {\bf 93.0} & 97.1 & {\bf 94.1} & 96.6 & 90.5 & 94.9 \\
{\bf Ours: Double-view} & {\bf 99.3} & 90.8 & {\bf 98.3} & 93.0 & {\bf 98.0} & {\bf 98.8} & {\bf 96.3} \\
\hline
\end{tabular}
\vspace{-3mm}
\end{table}
\end{savenotes}

\subsection{Kinship Verification \& Identification}
As before, we utilize the HSV 12-dim low-level features. In the experiments, we denote father, mother, son, and daughter as F, M, S, and D, respectively. Following \cite{lu2014neighborhood}, we measure the verification performance with the {\em verification rate}, defined by the number of correctly classified face pairs divided by the total number of face pairs in the test set. For identification, CMC curves are also used. We only use double-view training in this task since the information captured by parent-offspring pairs are more important.


Kinship verification between two views (one parent and one offspring) is the conventional setting, where we test our method on two datasets, \ie KinFaceW-I \cite{lu2014neighborhood} and KinFaceW-II \cite{lu2014neighborhood}. The former consists of 156 FS, 134 FD, 116 MS and 127 MD pairs, while the latter contains 250 pairs of each kin relation. The main difference between the two datasets is that each pair of face images in KinFaceW-II comes from the same photo while the image pairs in KinFace-I come from different photos. We follow the same protocol as that in \cite{lu2014neighborhood, dehghan2014look, qin2015tri} and use a 5-fold cross validation with balanced positive and negative pairs on the default training/testing split. Results are listed in Table \ref{tab:KinFaceW}.

On KinFaceW-II, our method significantly outperforms the competitors, but on KinFaceW-I ours is slightly worse. Our reasoning is that our current visual word representation using simple K-Means does not account for significant visual ambiguity in appearance when imaging factors (\eg lighting conditions, illumination, \etc) change substantially. This leads to large intra-cluster variations in visual words that our method does not currently handle well. To further investigate the different performances on both datasets, we use a smaller training set randomly sampled on KinFaceW-II such that it has the same size as KinFaceW-I, while keeping the same test set and record the results as ``reduced training set''. The results become slightly worse than the original training set, while still outperform other methods. These relatively good results, along with the worse results on KinFaceW-I, demonstrate that the size of training data is indeed important, but less important than the data sources.

\begin{table}\footnotesize
\caption{\footnotesize{Verification rate comparison (\%) on KinFaceW}}
\label{tab:KinFaceW}
\centering
\begin{tabular}{|l|l l l l l|}
\hline 
• & FS & FD & MS & MD & Mean \\ 
\hline 
\hline
 & \multicolumn{5}{c|}{KinFaceW-I} \\
\hline
Dehghan \etal \cite{dehghan2014look}  & 76.4 & 72.5 & 71.9 & 77.3 & 74.5 \\ 
Lu \etal \cite{lu2014neighborhood} & 72.5 & 66.5 & 66.2 & 72.0 & 69.9 \\ 
Qin \etal \cite{qin2015tri} & \textbf{76.8} & \textbf{76.8} & \textbf{74.6} & \textbf{78.0} & \textbf{76.6} \\ 
\hline 
\textbf{Ours}  & 63.5 & 65.0 & 63.8 & 75.6 & 67.0 \\ 
\hline 
\hline
 & \multicolumn{5}{c|}{KinFaceW-II} \\
\hline
Dehghan \etal \cite{dehghan2014look} & 83.9 & 76.7 & 83.4 & 84.8 & 82.2 \\ 
Lu \etal \cite{lu2014neighborhood} & 76.9 & 74.3 & 77.4 & 77.6 & 76.5 \\ 
Qin \etal \cite{qin2015tri} & 84.6 & 77.0 & 84.4 & 85.4 & 82.9 \\ 
\hline 
\textbf{Ours} & \textbf{85.4} & \textbf{81.8} & \textbf{86.6} & \textbf{90.0} & \textbf{86.0} \\ 
Ours (reduced training set) & 84.4 & 78.2 & 84.6 & 87.8 & 83.8 \\

\hline 
\end{tabular} 
\vspace{-3mm}
\end{table}


Next we use TSKinFace dataset \cite{qin2015tri} for three-view kinship verification (\ie father, mother, offspring), which contains 513 FM-S and 502 FM-D groups. Following \cite{qin2015tri}, we carry out a 5-fold cross validation with balanced positive and negative samples , and list the results in Table \ref{tab:TSKinFace}. As we see, our method performs consistently better than \cite{qin2015tri}.

\begin{table}\footnotesize
\caption{\footnotesize{Verification rate comparison (\%) on TSKinFace.}}
\label{tab:TSKinFace}
\centering
\begin{tabular}{|l|l l l l l l|}
\hline 
 & FS & FD & MS & MD & FM-S & FM-D \\ 
\hline 
Dehghan \etal \cite{dehghan2014look} & 79.9 & 74.2 & 78.5 & 76.3 & 81.9 & 79.6\\ 
Fang \etal \cite{fang2013kinship} &  69.1 & 66.8 & 68.7 & 67.9 & 71.6 & 69.8 \\
Lu \etal \cite{lu2014neighborhood} & 74.8 & 70.0 & 72.2 & 71.3 & 77.0 & 71.4\\ 
Qin \etal \cite{qin2015tri} & 83.0 & 80.5 & 82.8 & 81.1 & 86.4 & 84.4\\ 
\hline
\textbf{Ours} & \textbf{88.5} & \textbf{87.0} & \textbf{87.9} & \textbf{87.8} & \textbf{90.6} & \textbf{89.0} \\ 
\hline 
\end{tabular} \vspace{-3mm}
\end{table}

\begin{figure}[t]\footnotesize
\centerline{\includegraphics[width=0.75\columnwidth]{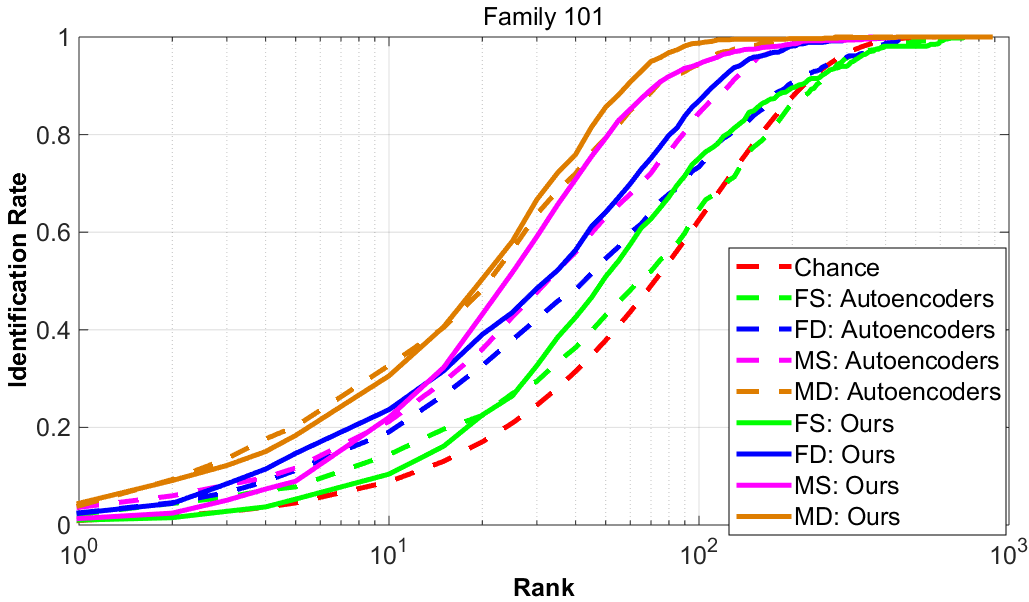}}
\caption{\footnotesize{CMC curve comparison with \cite{dehghan2014look} on Family 101.}}\label{fig:Family101}
\vspace{-4mm}
\end{figure}

Finally we employ the Family 101 dataset \cite{fang2013kinship} to investigate kinship identification, namely, identifying the correct parent/child among a set of candidates given one child/parent image. This dataset contains 14816 images that form 206 nuclear families belonging to 101 unique family trees. Following \cite{dehghan2014look}, we adopt 101 nuclear families and use 50 families for training and 51 families for testing. For each of the four kin relations, we train a model and use the model to match offspring images to all possible parent images. The CMC curves\footnote{We use the author's code (\url{http:// enriquegortiz.com/publications/FamResemblance.zip}) to produce the results.} are shown in Fig. \ref{fig:Family101} , and Table \ref{tab:Family101} lists the Area Under Curve (AUC) measure of the CMC curves.

\begin{table}\footnotesize
\caption{\footnotesize{AUC comparison (\%) on Family 101.}}
\label{tab:Family101}
\centering
\begin{tabular}{|l|l l l l l|}
\hline 
• & FS & FD & MS & MD & Mean \\ 
\hline 
Dehghan \etal \cite{dehghan2014look} & 88.8 & 91.3 & 94.3 & 96.4 & 92.7 \\ 
\hline 
\textbf{Ours} & \textbf{90.3} & \textbf{94.6} & \textbf{96.0} & \textbf{97.0} & \textbf{94.5} \\ 
\hline 
\end{tabular} \vspace{-3mm}
\end{table}

\subsection{Storage \& Computational Time}\label{ssec:time}
Storage ($S_t$ for short) and computational time during testing are two critical issues in real-world applications. In our method, we only need to store a feature matrix for each entity based on Eq. \ref{eqn:sp}, which is used to calculate similarities between different entities. The computational time can be roughly divided into two parts: (1) computing feature matrices $T_1$, and (2) predicting group membership $T_2$. We do not consider the time for generating low-level features, since different implementations vary significantly.

We record the storage and computational time using 300 visual words for both probe and gallery sets on VIPeR (two views), WARD (three views), and RAiD (four views). The rest of the parameters are the same as described in Section \ref{ssec:reid}. As we see, the storage per data sample and computational time are linearly proportional to the size of images and number of visual words. Our implementation is based on unoptimized MATLAB code\footnote{Our code is available at \url{https://zimingzhang.wordpress.com/source-code/}.}. Numbers are listed in Table \ref{tab:time}, including the time for saving and loading features. Our experiments were all run on a multi-thread CPU (Xeon E5-2696 v2) with a GPU (GTX TITAN). The method runs efficiently with very low demand for storage.

\begin{savenotes}
\begin{table}[t]\centering\footnotesize
\caption{\footnotesize{Average storage and computational time for our mdthod.}}\label{tab:time}
\vspace{1mm}
\begin{tabular}{|l||lll|}
\hline
& $S_t$ (Kb) & $T_1$ (ms) & $T_2$ (ms) \\
\hline
VIPeR & 110.7 & 52.9 & 0.6  \\
WARD & 113.7 & 99.7 & 1.5 \\
RAiD & 166.5 & 68.7 & 0.5 \\ 
\hline
\end{tabular}
\vspace{-3mm}
\end{table}
\end{savenotes}

%
%
%
%
%

\section{Conclusion}\label{sec:con}


In this paper, we propose a general parametric probability model for the group membership prediction (GMP) problem. We introduce the notions of view-specific and view-shared latent variables to capture visual information and commonality for each view. Using these two variables, we can factorize the group membership score into a tensor product, and thus propose a new visual word co-occurrence tensor feature to represent groups of data samples. In our parametric probability model, we can handle the multiple instance cases as well. Further we propose discriminatively learning a bilinear classifier for GMP, with the decision function as the marginalization over all latent variables. Our experiments on multi-camera person re-id and kinship verification tasks demonstrate the good predictive ability and computational efficiency of our method. As future work, we would like to explore other applications for our method such as activity retrieval \cite{Castanon_mm15}, and develop new approaches such as zero-shot recognition \cite{zhang_iccv15_ZS} and structured learning \cite{zhang2014person} for our problem.

\section*{Acknowledgement}
We thank the anonymous reviewers for their very useful comments. This material is based upon work supported in part by the U.S. Department of Homeland Security, Science and Technology Directorate, Office of University Programs, under Grant Award 2013-ST-061-ED0001, by ONR Grant 50202168 and US AF contract FA8650-14-C-1728. The views and conclusions contained in this document are those of the authors and should not be interpreted as necessarily representing the social policies, either expressed or implied, of the U.S. DHS, ONR or AF.

{\small
\bibliographystyle{ieee}
\bibliography{egbib-rev}
}

\end{document}